\documentclass{article}

\PassOptionsToPackage{numbers, compress}{natbib}

\usepackage[preprint]{tsrml_2022}

\usepackage{nicefrac}       
\usepackage{microtype}      

\usepackage{amsmath,amssymb,amsfonts}
\usepackage{graphicx}
\usepackage{xcolor}

\usepackage[T1]{fontenc}
\usepackage{xcolor}

\usepackage{newtxmath,newtxtext}
\usepackage{setspace}
\usepackage{lipsum}
\usepackage{fancyhdr}
\usepackage{url}
\usepackage{todo}
\usepackage{graphicx}
\usepackage{booktabs,totpages}

\usepackage{braket}

\usepackage[tikz]{bclogo}

\usepackage{pdfpages}
\usepackage{pgfgantt}
\usepackage{float}
\usepackage{wrapfig}
\usepackage{csquotes}

\usepackage{amsmath}
\newcommand*\diff{\mathop{}\!\mathrm{d}}

\usepackage{arydshln}
\usepackage{tikz}

\usepackage{amsthm}
\usepackage{amsmath}
\usepackage{bm}
\usepackage{thmtools,thm-restate}
\usepackage{enumitem}
\usepackage{pifont}
\usepackage{array}
\usepackage{float}
\usepackage{color}
\usepackage{multirow}
\usepackage{booktabs}
\usepackage{graphicx}
\usepackage{colortbl}
\usepackage{pythonhighlight}
\usepackage{hyperref}
\usepackage{comment}
\usepackage{lipsum}
\usepackage{bbm}
\usepackage{algorithm}
\usepackage[noend]{algpseudocode}
\usepackage{hyperref}
\usepackage{xcolor}
\usepackage{anyfontsize}
\usepackage{makecell}
\usepackage{mathtools}
\usepackage[T1]{fontenc}

\DeclareMathOperator{\Advantage}{Advantage}
\DeclareMathOperator{\WSD}{WSD}
\DeclareMathOperator{\VSD}{VSD}

\DeclareMathOperator{\Preference}{Preference}
\DeclareMathOperator{\NP}{NP}

\hypersetup{
colorlinks=true,
linkcolor=cyan,
citecolor=red,
urlcolor=blue}

\newcommand{\pluseq}{\mathrel{{+}{=}}}

\newcolumntype{P}[1]{>{\centering\arraybackslash}p{#1}}
\newcommand{\xingzhiswallow}[1]{{}}
\renewcommand{\paragraph}[1]{\textbf{\noindent{#1}}}

\newcommand*\colourcheck[1]{%
  \expandafter\newcommand\csname #1check\endcsname{\textcolor{#1}{\ding{52}}}%
}
\newcommand*\colourxmark[1]{%
  \expandafter\newcommand\csname #1xmark\endcsname{\textcolor{#1}{\ding{55}}}%
}
\colourcheck{blue}
\colourxmark{red}
\AtBeginDocument{%
  \providecommand\BibTeX{{%
    \normalfont B\kern-0.5em{\scshape i\kern-0.25em b}\kern-0.8em\TeX}}}

\makeatletter
\newcommand{\checkheight}[1]{%
  \par \penalty-100\begingroup%
  \setbox8=\hbox{#1}%
  \setlength{\dimen@}{\ht8}%
  \dimen@ii\pagegoal \advance\dimen@ii-\pagetotal
  \ifdim \dimen@>\dimen@ii
    \break
  \fi\endgroup}
\makeatother

\newcommand{\tup}[1]
{
 \relax\ifmmode
 \langle #1 \rangle
 \else $\langle$ #1 $\rangle$ \fi
}

\newcommand{\swallow}[1]{ }

\begin{document}
\title{Provable Fairness for Neural Network Models \\
using Formal Verification}
\author{%
Giorgian Borca-Tasciuc  \quad Xingzhi Guo \quad Stanley Bak \quad Steven Skiena  \\
Department of Computer Science, Stony Brook University, USA\\
\texttt{\{gborcatasciu,xingzguo,sbak,skiena\}@cs.stonybrook.edu}
}



\maketitle

\thispagestyle{fancy}

\begin{abstract}

\noindent
Machine learning models are increasingly deployed for critical decision-making tasks, making it important to verify that they do not contain gender or racial biases picked up from training data.
Typical approaches to achieve fairness revolve around efforts to clean or curate training data, with post-hoc statistical evaluation of the fairness of the model on evaluation data.
%
%
In contrast, we propose techniques to \emph{prove} fairness using recently developed formal methods that verify properties of neural network models.
Beyond the strength of guarantee implied by a formal proof, our methods have the advantage that we do not need explicit training or evaluation data (which is often proprietary) in order to analyze a given trained model.
In experiments on two familiar datasets in the fairness literature (COMPAS and ADULTS), we show that through proper training, we can reduce unfairness by an average of 65.4\% at a cost of less than 1\% in AUC score.

\end{abstract}

\section{Introduction}

Machine learning models are increasingly deployed for critical decision-making tasks (e.g., Bail decision\cite{angwin2016machine} ). It is important to verify that they do not contain gender or racial biases picked up from training data.   
Typical approaches to achieve fairness revolve around efforts to curate training data, with post-hoc fairness evaluation on testing data (which is often proprietary).
In contrast, we propose techniques to \emph{prove} fairness using formal methods that verify properties of neural network models.  
Beyond the strength of guarantee implied by a formal proof, our methods have the advantage that we do not need explicit training or testing data to analyze a given trained model.
Formal approaches to neural network verification have been developed over the past few years~\cite{liu2019algorithms,tran2020verification,katz2017reluplex,gehr2018ai2} that can in certain cases prove properties over all possible executions of a specific network.
Existing work has focused on networks used in safety-critical control algorithms and robotics~\cite{ivanov2019verisig,xiang2020reachable}, or guaranteed robustness for visual perception networks~\cite{tran2019safety}.

In this paper, we apply such methods to the task of validating fairness properties of classification and regression models.
We systematically explore classes of fairness specifications that can be evaluated statistically as well as using our developed formal approaches.
We consider using formal methods to analyze fairness for \emph{classification tasks with explicit input labels}. Here one of the input fields for each record explicitly encodes a sensitive property, such as race, gender, or age. 
Informally, fairness dictates that a given model $M$ should perform ``the same'' over all possible values of these sensitive fields.
Note that excluding such inputs when training a model does not solve the fairness problem, as sensitive properties are often inferable from other input variables (e.g. the zip code of an African-American neighborhood).
    
\begin{figure*}[t]
    \centering
    \includegraphics[width=1.0\columnwidth]{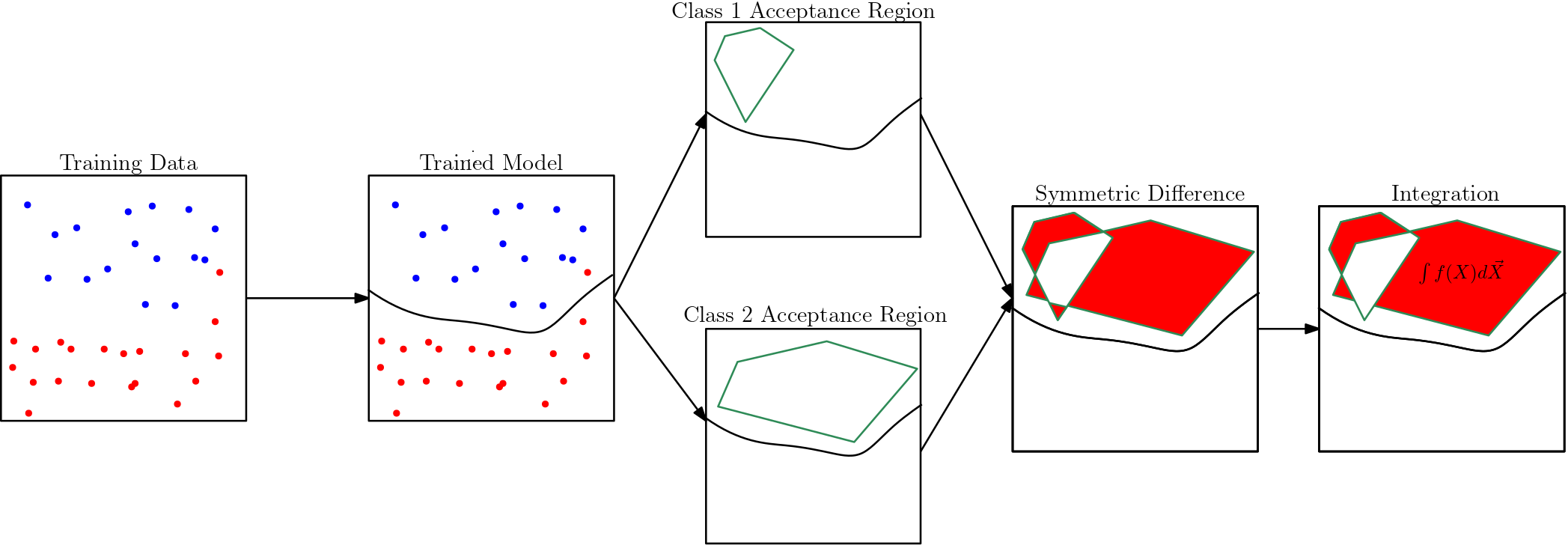}
    \caption{In our verification approach, each partition $\Theta$ of the network input space defines a domain for integration over a given input probability density function $P$, calculating a probability of the input space with a fixed output classification. 
    By repeating and summing over all partitions, we can evaluate fairness metrics for the network.}
    \vspace{-3mm}
    \label{fig:input_split}
\end{figure*}

Formal analysis approaches are most often applied to fully connected networks with ReLU activation functions~\cite{vnncomp2021}.
For such networks, a binary decision model partitions the space of possible inputs into a collection of geometric polytopes, such that all points within a particular polytope are labeled homogeneously as all positive or all negative. 
The fairness properties which we are concerned with revolve around showing that these geometric regions are comparable for two distinct labels or groups, as illustrated in Figure~\ref{fig:input_split}.
There are several possible provable properties depending upon how we define that $M$ performs “the same” for different groups, including:  

\begin{itemize}


\item {\em Volumetric symmetric difference} ---
Here, we say that $M$ is provably fair if the volume of the non-overlapping acceptance regions for the protected classes is below a certain threshold.
\item {\em Probability-weighted symmetric difference} ---
Here we say that $M$ is provably fair with respect to input distributions $A$ and $B$ if the percentage of each population which would change classified label by changing label of the protected class is below a certain threshold.
\item{\em Net Preference} --- Here we say that $M$ is provably fair with respect to input distributions $A$ and $B$ if there is not a large difference in acceptance probability if the criteria used to accept the population from distribution $A$ is used evaluated on the population from distribution $B$.

\end{itemize}

Our major contributions in this paper are:

\begin{itemize}
    \item {\em Provable Fairness Guarantees \textbf{without} Training or Evaluation Data} -- We define a class of fairness guarantees which can be formally computed on neural classification models for real tasks.
    A particularly exciting aspect of our approach is that verification requires only black box access to the trained model.  Previous approaches to fairness evaluation require access to
    training or evaluation data, usually deemed confidential and proprietary for consumer product models.  One of the metrics we introduce require no training and evaluation data, nor knowledge of the distribution of the class features. The other two metrics we introduce only require knowledge of the probability distribution of the class features in the applied setting, which may be estimated either from the training and evaluation data or using external empirical studies demographic in nature.

    \item {\em Addressing Scaling Issues} -- Verifying the properties of neural networks requires sophisticated geometric algorithms in high-dimensional spaces.
    The number and complexity of polytopes defining our analysis grow exponentially with the size of the model, and its parameter space.
    Still, the state-of-the-art in formal verification has advanced steadily despite theoretical intractability~\cite{katz2017reluplex}.
    We identify the bottleneck in our approach as the volume computation step and propose alternate algorithms for volume integration that can further push the scalability limits of our approach.
    
\end{itemize}

\xingzhiswallow{
This paper is organized as follows.

Section \ref{sec:prior-work} reviews previous work in formal verification techniques for neural networks, and also in the field of algorithmic fairness.
We discuss our formal verification methods in Section \ref{sec:methods}.
Our experimental results demonstrating the trade off between model quality and quantified fairness are presented in Section \ref{sec:experimental-results}.
Finally, we do a critical performance analysis on the scalability of our methods in Section \ref{sec:scalability}, pointing to future research directions in the verification of model fairness. 
Our source code and datasets will be released upon publication, and included in this submission at \textcolor{blue}{\url{https://bit.ly/3MC64Ke}}
}

\section{Related Work}
\label{sec:prior-work}

\paragraph{Formal verification methods for neural networks}: The formal verification methods~\cite{xiang2018verification,liu2019algorithms,albarghouthi-book} perform set-based analysis of the network, rather than executing a network for individual input samples.
Given a neural network $f_{NN}: \mathbb{R}^n \rightarrow \mathbb{R}^m$, an input set $X \subseteq \mathbb{R}^{n}$, and an unsafe set of outputs  $U \subseteq \mathbb{R}^{m}$, the \emph{open-loop NN verification problem} is to prove that for all allowed inputs $x \in X$, the network output is never in the unsafe set, $f_{NN}(x) \notin U$.
One way to solve the verification problem is to compute the range of the neural network function for a specific input domain $X$, and then check if the range intersects the unsafe states $U$~\cite{bak2020cav, vincent2021reachable}.
In this work, we reuse this range computation approach in order to evaluate how probability distributions propagate through the networks.
%
%
%
%
Technically, two core operations are performed in our work: (1) computing the range of the neural network as a union of polytopes, and  (2) computing the volume of the output polytopes. Differing from the formal verification methods used for safety and motion planning, we explore the possibility to apply it for neural network fairness measure. 
%

\paragraph{Fairness in neural networks}: Machine learning fairness has also received increasing attention recently~\cite{mehrabi2021survey,caton2020fairness,bellamy2018ai}.
Existing work on fairness typically uses the input data and samples the network to provide estimates of fairness~\cite{bastani2019probabilistic}. However, such testing samples may not cover whole population (e.g., biased testing samples), and thus cannot guarantee the model fairness.
In contrast, our techniques use external distributions over input data and perform set-based range computation of neural networks, without the need for any testing data for fairness evaluation.
%
%
A limited amount of existing research exists on provable fairness.
One framework focuses on proving \emph{dependency
fairness}~\cite{urban2020perfectly,galhotra2017fairness}, which strives to prove the outputs are not affected by certain input features.
This method is based on forward and a backward static analysis as well as input feature partitioning.
Another recent approach~\cite{ruoss2020learning} focuses on \emph{individual fairness}, which essentially means that similar individuals get similar treatments.
In contrast, our provable fairness metrics are defined over geometric properties of the network outputs, such as the symmetric difference between the ranges with different values of sensitive inputs. 

\section{Methods: Provable Fairness using Formal Verification}
\label{sec:methods}

\subsection{Acceptance Region}
We evaluate the fairness of the models by looking carefully at the exact acceptance regions of the models. 
Let $f(\vec{x})$ be the label assigned by the neural network for the value $\vec{x}$ in the input space. The acceptance region for class C, which we denote as $\rho(C)$ is the region in the input space for which $\vec{x}\in C\land f(\vec{x})=\ell$. The label $\ell$ that defines the acceptance region is user-definable based on the properties of the neural network they are interested in exploring. We denote the acceptance region of a class $C$ as $\rho(C)$.
\subsection{Proposed Metric}
\label{sec:metric}
We next introduce the metrics that we use to evaluate the fairness of our models. These metrics serve as quantifications of the legal notions of disparate treatment and disparate impact \cite{legaldef}.
The weighted symmetric difference (WSD) quantifies the total difference in the model’s behaviour towards different classes. The WSD quantifies disparate treatment.
It is weighted in order to discount differences in behaviour for unrealistic sets of features that are improbable to ever appear in the input. Formally, the metric is defined as:
\begin{align*}
\WSD(C_1, C_2) &= \Advantage(C_1, C_2) + \Advantage(C_2, C_1) \\
\Advantage(C_1, C_2) &= \int_{\rho(C_1)} P(\vec{X}|C_1)d\vec{X} - \int_{\rho(C_1)\cap\rho(C_2)} P(\vec{X}|C_2)d\vec{X} \\
\end{align*}
The $\Advantage(C_1, C_2)$ metric quantifies how many examples from class $C_1$ would have not been in the acceptance region of the model if they were of class $C_2$, with all other features held equal. It is a special case of the discrimination score~\cite{discscore}. 

The volumetric symmetric difference (VSD) also quantifies the total difference in the model’s behaviour towards different classes. It may be used when information about the input feature distribution for the two classes is unavailable. The VSD also quantifies disparate treatment.
\begin{align*}
\VSD(C_1, C_2) &= |\rho(C_1) - \rho(C_2)| + |\rho(C_2) - \rho(C_1)| \\
|\rho(C_1) - \rho(C_2)| &= \int_{\rho(C_1)} 1\cdot d\vec{X} - \int_{\rho(C_1)\cap\rho(C_2)} 1\cdot d\vec{X} \\
\end{align*}

Finally, the net preference (NP) quantifies how much the model prefers the features of one class over the other when assigning a label. The NP quantifies disparate impact. It is useful for investigating whether the model makes its decisions on variables strongly correlated with the protected class. When evaluating the preference, the acceptance region of $C_1$ is used in order to discount possible differences in the acceptance regions, which can be investigated using the $\WSD$ and the $\VSD$. Net preference is similar to that disparate impact metric given by Feldman et. al\cite{feldman}, with one key advantage: by taking into account the possible difference in acceptance regions between the two classes, NP focuses on capturing disparity that arises from criteria that a priori might seem agnostic to the sensitive class. Feldman et al's metric captures disparities that arise both from disparate treatment and from disparate impact, making it difficult to further investigate the source of the problem.
\begin{align*}
\NP(C_1, C_2) &= \max\{|\Preference(C_1, C_2)|, |\Preference(C_2, C_1)|\} \\
\Preference(C_1, C_2) &= \int_{\rho(C_1)} P(\vec{X}|C_1)d\vec{X} - \int_{\rho(C_2)} P(\vec{X}|C_2)d\vec{X} \\
\end{align*}
The preference function here quantifies the difference in the acceptance probability of the two classes when evaluated by the acceptance criteria of the first class.

\subsection{Verification Approach}

Our technical approach to verification is based on set-based execution of neural networks, extended to include probability distributions.
Given a set of possible inputs, we propagate the set through the network to see the range of the network, the possible outputs.
%
%
In our approach, we use the linear star set representation~\cite{duggirala2016parsimonious,tran2020cav} to propagate sets of states through the network.
%
%
%
%
After analysis, the entire input set is partitioned into a collection of polytopes that map to outputs with an identical label.

This representation of sets of inputs allows for defining a simple procedure for verifying the fairness of a neural network. There is an assumption that it is possible to encode $C_1$ and $C_2$ directly into the input features by fixing one or more of the features in the input to the appropriate values. The procedure is as follows:
\begin{enumerate}
    \small
    \item Use neural-network reachability analysis in order to calculate the acceptance regions for each of the classes: $\rho(C1)$ and $\rho(C2)$, then calculate $\rho(C_1)\cap \rho(C_2)$.
    \item Calculate $\WSD(C_1, C_2)$ using the probability distributions $P(\vec{X}|C1)$ and $P(\vec{X}|C_2)$ over the data.
\end{enumerate}

The calculation of $\WSD(C_1, C_2)$ requires a probability distribution over the data. 
For our experiments, we estimate the probability distribution from the input data, although external demographic data can be used to create these distributions if input data is not available.
The details of this method are in the appendix.

The neural network reachability analysis returns two lists of polytopes, encoding the acceptance region of each class: $\rho(C_1)=\{A_1, A_2, A_3, \ldots A_N\}$ and $\rho(C_2) = \{B_1, B_2, B_3,\ldots B_M\}$. 
Calculating the intersection between the two regions is thus a matter of calculating the non-empty intersection between all pairs of polytopes in set induced by the
Cartesian product $\rho(C_1) \times \rho(C_2)$. 
As the polytopes are
$\mathcal{H}$-polytopes, they are
specified as a list of constraints $C\alpha\leq d$. 
The constraints of two polytopes are simply concatenated to create the intersection of the two polytopes.
Linear programming can be used to determine if the intersection is empty.

\subsection{Integration over the Probability Distribution}
The final step for calculating the $\WSD(C_1, C_2)$ for a model requires the integration the probability distributions $P(\vec{X}|C_1)$ and $P(\vec{X}|C_2)$ over a union of polytopes. In our experiments, direct integration of the probability distribution over the polytope proved too slow once more then two dimensions needed to be integrated over. However,  determining the volume of the polytope has a better time complexity then integrating over the volume of the polytope.
Although we use \textsc{QHull} to find the volume due to its accessibility, algorithms that are linear in the number of vertices exist. We take advantage of this fact, computing the probability by discretizing the input space into a evenly-spaced grid. We calculate the intersection of each polytope with each region of the grid. We find the volume of this intersection and multiply it by the probability density at that point in the grid. This gives us a large speed-up in runtime for only a small cost in precision.

\section{Experimental Results}
\label{sec:experimental-results}


\xingzhiswallow{

To investigate the fairness of the neural networks, we train multiple models on three datasets that are commonly used in model fairness studies, and measure their fairness with our method according to the metrics in Section~\ref{sec:metric}. Note that our measurement does not require expensive data sampling or availability of testing data, making it different 
compared to other works as described in Section~\ref{sec:prior-work}.
Moreover, we also study two fairness-oriented data-augmentation strategies, effectively producing more \textit{fair} models at little expense of predictive power. 

In the following subsections, we present the statistics of the datasets and models, report both the performance and the fairness of the models trained by various data-augmentation strategies, including non-/fairness-enhanced.  
}

We train various models on three datasets that are commonly used in model fairness studies, and measure their fairness with our proposed metrics described in Section~\ref{sec:metric}. Note that our measurement does not require expensive data sampling or availability of testing data, making it different compared to other works as described in Section~\ref{sec:prior-work}.

\subsection{Dataset and Model Training }
We selecte three common datasets (COMPAS  \footnote{Correctional Offender Management Profiling for Alternative Sanctions: \url{https://www.propublica.org/article/how-we-analyzed-the-compas-recidivism-algorithm/}}, 
ADULTS  \cite{Dua:2019},
HEALTH  \footnote{\url{https://www.kaggle.com/c/hhp}}) 
that contains both categorical and continuous variables as the input features. 
All the datasets contains at least one protected feature (e.g., Race or Sex), which we consider as a potential factor that incurs unfair inductive bias during model training. 
In the following, we describe four training strategies 
\footnote{The details of strategies 3 and 4 can be found in Algorithm \ref{algo:fair-data-aug} in Appendix.} 
that increasingly alleviate such unfairness, demonstrating that our proposed metrics well capture the improvement in fairness. 
We present the details in Table \ref{tab:res-adults} and Table \ref{tab:dataset-stats} in Appendix due to space limitations

\xingzhiswallow{
    
    \begin{table*}[ht]
       \centering
       \label{tab:dataset-stats}
       \caption{The statistics of dataset. For each dataset we split data into train/val/test set with the ratio 70\%/15\%/15\%, respectively. For each continuous feature, we applied min-max normalization. For simplification, we use binary label for the protected feature: Race (White v.s. African American) and Sex (Male v.s. Female).}
        \begin{tabular}{p{0.18\textwidth} p{0.10\textwidth} p{0.1\textwidth} p{0.15\textwidth}}
        \toprule
         \textbf{Dataset} & \textbf{COMPAS}   & \textbf{ADULTS} & \textbf{HEALTH} \\
        \midrule 
         \#Samples & 2,363  & 43,031  & 124,086 \\
         \#Categorical Feat. & 2 & 5 & 11  \\
         \#Continuous Feat. & 2  & 3 & 11 \\
         Protected Feat. & Race & Race & Sex   \\
         \#Classes & Binary  & Binary  & Binary \\
         \bottomrule
        \end{tabular}
    \end{table*}

}


\begin{itemize}
\item[1.] {\em  Baseline}:  Use the original data to train the model, without any augmentation or filtering.

\item[2.] {\em Protected Feature Permuted}: Randomly shuffle the protected feature (Race or Sex) in training data, then train the model once. The approach breaks the correlation between the feature and the label, for example, making prediction class-agnostic.

\item[3.] {\em Data-Removed}: Re-train the model with the \textbf{removed} data, iteratively deleting the training data which may cross the decision boundary (threshold 0.5) when the protected feature is flipped. This is a simplified version of the technique presented in \cite{dataremoval}, without employing the ranking algorithm. 

\item[4.] {\em Data-Augmented}: Re-train the model with the \textbf{augmented} data, iteratively creating training data which may cross the decision boundary (threshold 0.5) when the protected feature is flipped. The max number of training data is limited to three times of the original data size. This is a combination of techniques applied in~\cite{dataaug1} and~\cite{dataaug2}.
\end{itemize}



\subsection{Results}

\xingzhiswallow{
There are several key results. The first is that without any measures to ensure fairness during the training process, the trained models demonstrate a significant degree of unfairness. Inspecting the WSD of various models trained on the COMPAS dataset in Table~\ref{tab:res-compas}, there is a significant number of individuals for whom the model decision would have switched had only their race been different. This demonstrates the importance of taking measures to ensure model fairness, especially in domains that have historically been race-sensitive. 
}
\paragraph{The baseline models are unfair without fairness training}: Inspecting the WSD of various models trained on the COMPAS dataset in Figure~\ref{fig:auc-vs-fairness-compas}, there are a significant number of individuals for whom the model decision would have switched had only their race been different.   Figure \ref{fig:auc-vs-fairness-adults-long} and Table \ref{tab:res-adults} in Appendix show the consistent pattern in all trials on two datasets.

\begin{figure}[ht]
    \centering
    \includegraphics[width=0.9\textwidth]{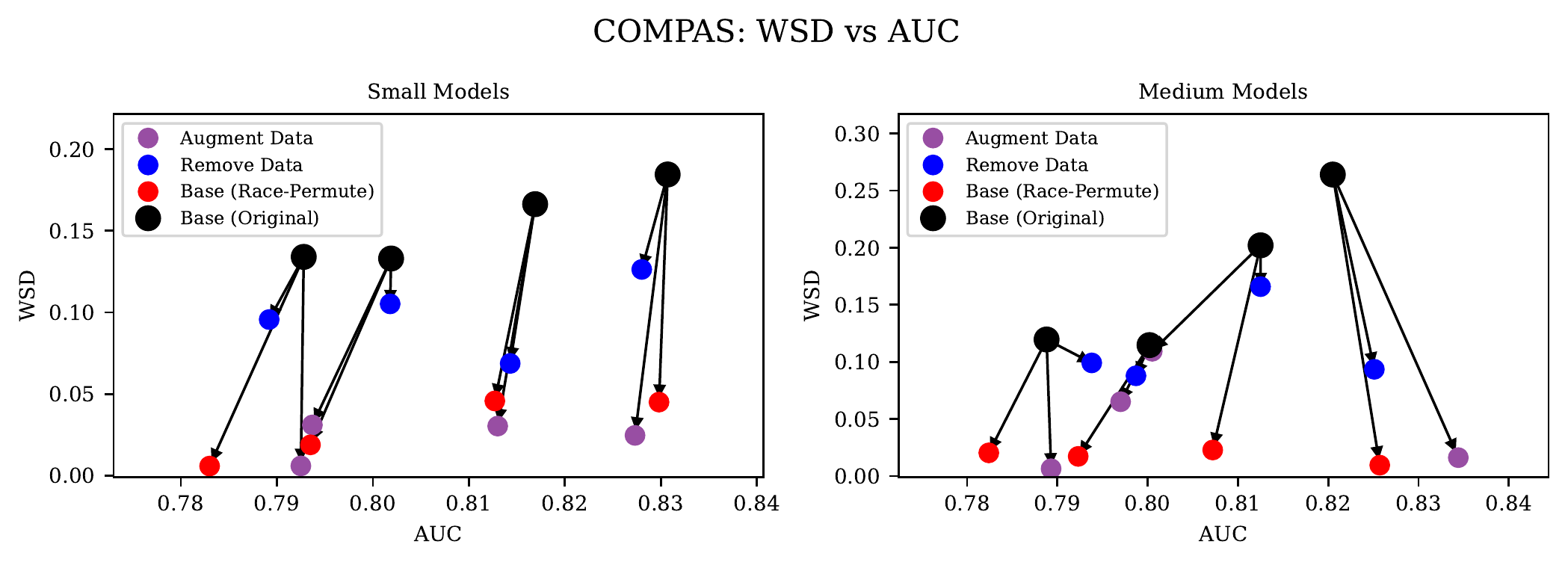}
    \vspace{-3mm}
    
        \caption{ COMPAS Dataset: Model performance/AUC vs Model fairness/NP under three fairness-sensitive training methods.  Arrows down and to the left reflect models that are fairer but less accurate than the original mode.  We obtain large improvements in fairness at little cost in accuracy. }
    \vspace{-3mm}
    \label{fig:auc-vs-fairness-compas}
\end{figure}

\xingzhiswallow{
The WSD and NP of various models on the ADULTS dataset in Table~\ref{tab:res-adults} demonstrate another important result. In the presence of many features on which to base its decision, a model might have small WSD, indicating that there are few individuals for which had their race been different, the model decision would have been different. However, a well-known problem is the ability to use a \textit{proxy variable}, such as ZIP code and other correlated variables, in order to determine the race of an individual. This can be seen in the fact that the WSD is small but the NP is large for the original models in Table~\ref{tab:res-adults}, indicating that the model has a strong preference for features correlated with a certain race.  
}

\xingzhiswallow{
Comparisons of the $\WSD$ and the $\VSD$ show that when the probability distribution on the input features is not available, the $\VSD$ is a suitable proxy for the $\WSD$. Reductions in the $\WSD$ are accompanied by similar reductions in the $\VSD$. In 11 out of the 16 trials, the model with the smallest $\WSD$ was also the model with the smallest $\VSD$.
}
\paragraph{$\VSD$ is a suitable proxy for the $\WSD$ when the probability distribution on the input features is not available}: Reductions in the $\WSD$ are accompanied by similar reductions in the $\VSD$. In 11 out of the 16 trials, the model with the smallest $\WSD$ was also the model with the smallest $\VSD$.

\xingzhiswallow{
Finally, our results demonstrate the effectiveness of our data-augmentation method in order to improve fairness. The data-augmentation method leads to a significant improvement in the fairness of the models, with only a small drop in model accuracy (and sometimes an improvement). The data-augmentation method is often able to significantly outperform or achieve parity with the baseline of permuting the race in many cases for the WSD metric. However, unlike the race-permutation method, the method is also effective in greatly reducing the NP. Thus, the models rely much less on proxy variables for race. 
}

\paragraph{The trade-off between fairness and accuracy}: The data-augmentation method leads to a significant improvement in the fairness of the models, with only a small drop in model accuracy (and sometimes an improvement). The data-augmentation method is often able to significantly outperform or achieve parity with the baseline of permuting the race in many cases for the WSD metric. However, unlike the race-permutation method, the method is also effective in greatly reducing the NP. The data-augmented models rely much less on proxy variables for race.

\xingzhiswallow{
\begin{figure*}
    \centering
    \includegraphics[page=1,width=0.95\textwidth]{figs/fig-compas-fairness-vs-auc-scatter-diff.pdf}
    \includegraphics[page=2,width=0.95\textwidth]{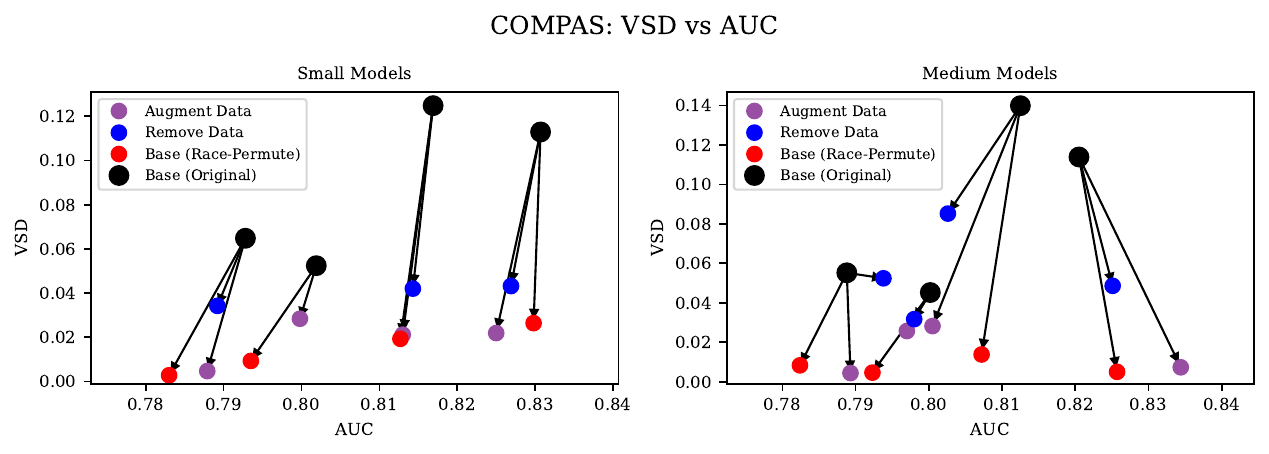}
    \includegraphics[page=3,width=0.95\textwidth]{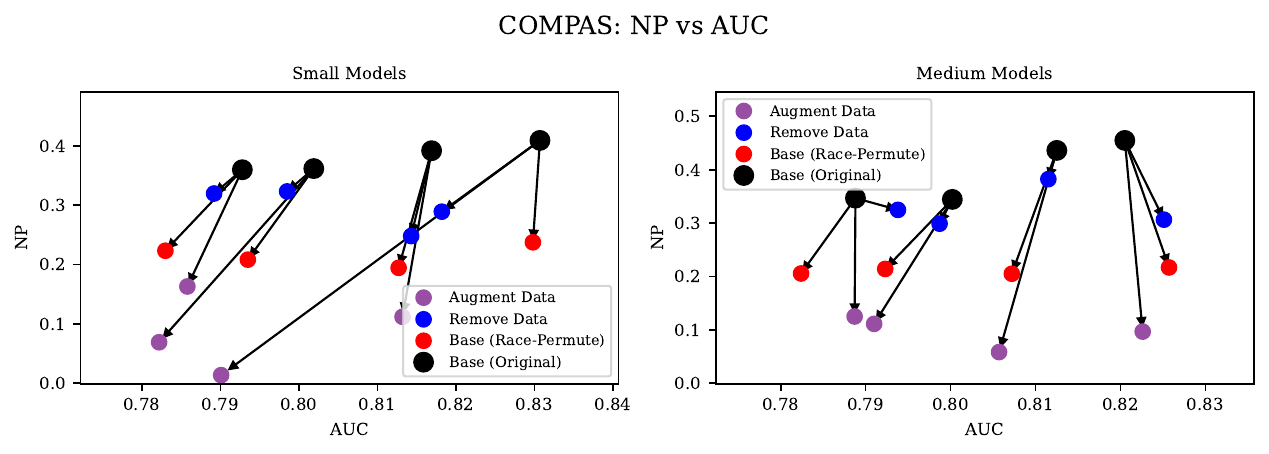}

    \caption{COMPAS Dataset: Model performance/AUC vs Model fairness/NP under three fairness-sensitive training methods.  Arrows down and to the left reflect models that are fairer but less accurate than the original mode.  We obtain large improvements in fairness at little cost in accuracy. }
    \label{fig:auc-vs-fairness-compas}
\end{figure*}

\begin{figure*}
    \centering
    \includegraphics[page=1,width=0.95\textwidth]{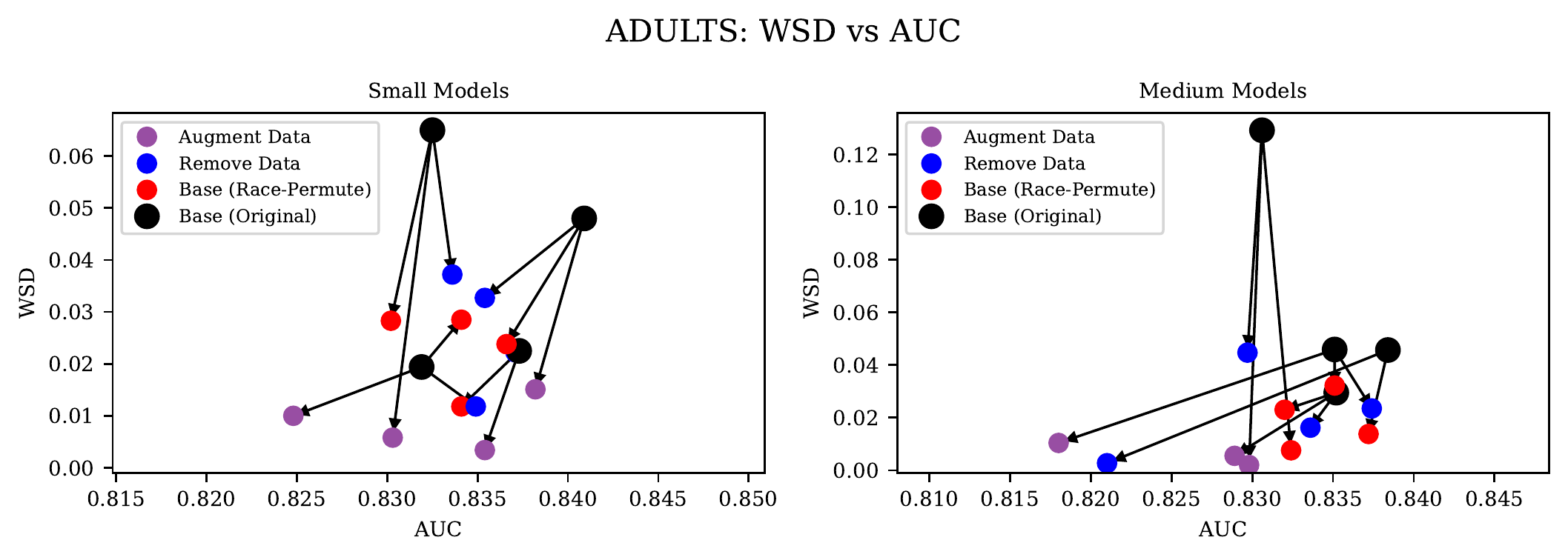}
    \includegraphics[page=2,width=0.95\textwidth]{figs/fig-adults-fairness-vs-auc-scatter-diff.pdf}
    \includegraphics[page=3,width=0.95\textwidth]{figs/fig-adults-fairness-vs-auc-scatter-diff.pdf}
    \caption{ADULTS Dataset: Model performance/AUC vs Model fairness/NP under three fairness-sensitive training methods.  Arrows down and to the left reflect models that are fairer but less accurate than the original mode.  We obtain large improvements in fairness at little cost in accuracy. }
    \label{fig:auc-vs-fairness-adults}
\end{figure*}

}

\subsection{Limitations of the Current Method}
\label{sec:scalability}

\xingzhiswallow{

\paragraph{Scalability}: The health dataset was omitted due to scalability issues of the current method. Performing the reachability analysis and identifiying the union of polytopes corresponding to the acceptance region of each of the classes took a total of less than two minutes ($\sim 100$s) for the 197 models tested for the health dataset. However, attempts to integrate the probability distribution over the probability distribution in a reasonable time frame failed. The integration method requires discretizing the input space, dividing each continuous dimension into $x$ divisions. If there are $d$ continuous dimensions, this creates $x^d$ total divisions, which are iterated over. As there are $11$ continuous dimensions in the health dataset, this generates $x^{11}$ total divisions. The precision of the integration is determined by the number of divisions. For a reasonably precise answer, at least $10$ divisions are necessary. With at least $10^{11}$ divisions per polytope, this naive integration method was impractical. 

Another approximation for the integral of the probability distribution over the polytope was to sample the probability distribution at the center of the polytope and scaling it by the volume of the polytope. Our method for finding the volume requires enumerating the polytope vertices. This becomes infeasible for $11$-dimensional polytopes as well.

There are alternative methods to finding the volume of the polytope that do not require
enumeration of the vertices. Emeris and Fisikopoulos~\cite{volapprox} give a polytope
volume approximation algorithm that takes polynomial time in the desired precision. As the number of vertices of a polytope can grow exponentially with the dimension, this improves the practicality of our method.
}

\paragraph{Integration}:
The procedure for discretizing the input space into a grid makes analysis infeasible for more than a small number of continuous inputs. 
If there are $11$ continuous variables, and each axis is broken into $10$ bins, this yields $10^{11}$ total regions. 
Thus, the number of continuous variables must be limited. 
There are workarounds to this approach. A direct discretization of the polytope based on the gradient of the $P(\vec{X}|C)$ would greatly reduce the number of regions while even allowing for improvements in precision. 
Additionally, efficient methods for integration of \textit{polynomials} over polytopes exist, which can be exploited if the $P(\vec{X}|C)$ takes the form of a polynomial.

\paragraph{Handling of One-Hot Features}:
The manner in which we handle one-hot features in the reachability analysis propagates every combination of the one-hot features through the neural network. This makes reachability analysis infeasible for more than a moderate number of one-hot features: for example, if there are $11$ one-hot features, each with $10$ different options, there are $10^{11}$ different combinations that must be handled.

\paragraph{Vertex Enumeration}:
Our method performs vertex enumeration by computing the convex-hull of the dual of the constraints. This is fast enough for the limited number of dimensions on which we try our method, as currently the limiting factor is the exponential growth of the size of the discretized grid. However, it still bears a significant time penalty and this method can be improved by using the Avis-Fukuda method cite{avis1991pivoting}, which runs in linear time on the number of vertices.

\paragraph{Polytope Volume Computation}:
\textsc{QHull} is used to compute the volume of the polytope due to its accessibility and robustness. However, this becomes suboptimal as the input dimensionality and the number of vertices grows. Lawrence's algorithm runs in linear time in the number of vertices, providing a large potential speed-up. 
%

\paragraph{Explicit Encoding of Classes}:
Our analysis methods assumes that the classes analyzed can be represented as polytopes in the input-space. This limits the applicability of the method to analyzing fairness of latent variables. However, if one has an estimate for the form of the latent variable, this can be approximated with a union of polytope in the input space and analyzed.

\vspace{-3mm}
\section{Conclusion}
We introduce a method to formally analyze the fairness of neural network models. Our fairness analysis methods do not require access to the model's training data, thus enabling model fairness evaluation in a broad range of contexts. We use our method to evaluate the fairness of several models, and demonstrate that models trained without intervention often demonstrate a significant degree of unfairness. We demonstrate that for circumstances in which the probability distribution on the inputs is not accessible, a probability-free approach provides a suitable substitute. 
Our results show that our data-augmentation method is a simple yet effective method for improving the model's fairness. Critically, it is also effective at reducing reliance on proxy variables of the protected class. 
We have explored the limits of scalability of our method: showing that there is still much room to improve runtime performance, by replacing the algorithms that enumerate polytope vertices and find the volume of the polytope from these vertices with more optimal ones. 
\label{sec:conclusion}

\bibliographystyle{abbrv}
\bibliography{refs}

\appendix
\section{Appendix}

\subsection{Dataset Detail}

\paragraph{COMPAS dataset} records the risk score of defendants based on his/her demographics and prior criminal history. We train the models to predict whether the individual's risk score is above or below 5 as a binary classification task. 

\paragraph{HEALTH dataset} \footnote{\url{https://www.kaggle.com/c/hhp}} contains the Charlson Index \footnote{Charlson Comorbidity Index predicts the ten-year mortality for a patient who may have a range of comorbid conditions} based on the patient's age, sex and prior medical history. The models are trained to predict whether the Charlson Index is above or below the median value.

\paragraph{ADULTS dataset} \cite{Dua:2019} contains the yearly income range (above or below \$50K) of the individuals together with demographics, education and occupational information. 

\begin{table*}[ht]
   \centering
   \label{tab:dataset-stats}
   \caption{The statistics of dataset. For each dataset we split data into train/val/test set with the ratio 70\%/15\%/15\%, respectively. For each continuous feature, we applied min-max normalization. For simplification, we use binary label for the protected feature: Race (White v.s. African American) and Sex (Male v.s. Female).}
    \begin{tabular}{p{0.18\textwidth} p{0.10\textwidth} p{0.1\textwidth} p{0.15\textwidth}}
    \toprule
     \textbf{Dataset} & \textbf{COMPAS}   & \textbf{ADULTS} & \textbf{HEALTH} \\
    \midrule 
     \#Samples & 2,363  & 43,031  & 124,086 \\
     \#Categorical Feat. & 2 & 5 & 11  \\
     \#Continuous Feat. & 2  & 3 & 11 \\
     Protected Feat. & Race & Race & Sex   \\
     \#Classes & Binary  & Binary  & Binary \\
     \bottomrule
    \end{tabular}
\end{table*}

\subsection{Model Detail}
We train the small and middle size models implemented in PyTorch. We use three 3 and 4 hidden layers, respectively, for small and middle models, with 8 neurons and ReLu activation per layer. 

\begin{table*}
\caption{
Model performance(\textit{AUC}) \textit{v.s} Model Fairness over 4 runs for the small/medium models on the ADULTS and COMPAS datasets.  Fairness-sensitive training methods (race permutation, and data removal/augmentation methods) reduce unfairness by an average of 65.1\% while reducing model quality (AUC) by less than 1\%.}
    \scriptsize

\centering
\begin{tabular}{lll|rrll|rrll}
 &  &  & \multicolumn{4}{c}{COMPAS}  &  \multicolumn{4}{|c}{ADULTS} \\
\toprule
\multirow{2}{4em}{Model Size}       & \multirow{2}{3em}{Trial ID}  & \multirow{2}{4em}{Measure}           &  \multirow{2}{4em}{Baseline} &  \multirow{2}{4em}{Perm Race} & \multirow{2}{4em}{Rem\\ Data} & \multirow{2}{4em}{Aug\\ Data} &  \multirow{2}{4em}{Baseline} &  \multirow{2}{4em}{Perm\\ Race} & \multirow{2}{4em}{Rem\\ Data} & \multirow{2}{4em}{Aug\\ Data} \\
\\
\midrule
Small  &  0  &  AUC  & \textbf{0.8169} &          0.8127  &                        0.8143  &                          0.8130 & \textbf{0.8373} &          0.8341  &                        0.8371  &                          0.8354 \\ 
&           &  WSD  &     0.1663  &          0.0457  &                        0.0687  & \textbf{0.0303} &     0.0225  &          0.0118  &                        0.0222  & \textbf{0.0034}\\
&           &  VSD  &     0.1248  & \textbf{0.0193} &                        0.0420  &                          0.0212 &     0.2042  & \textbf{0.0846} &                        0.2042  &                          0.1041 \\
&           &  NP  &     0.3919  &          0.1943  &                        0.2480  & \textbf{0.1118} &     0.1155  &          0.0945  &                        0.1155  & \textbf{0.0168}\\ \hline
&  1  &  AUC  & \textbf{0.7928} &          0.7830  &                        0.7892  &                          0.7925 &     0.8319  &          0.8341  & \textbf{0.8349} &                          0.8248 \\
&           &  WSD  &     0.1340  & \textbf{0.0058} &                        0.0956  &                          0.0059 &     0.0194  &          0.0285  &                        0.0118  & \textbf{0.0100}\\
&           &  VSD  &     0.0648  & \textbf{0.0028} &                        0.0342  &                          0.0047 &     0.2661  &          0.2338  & \textbf{0.1007} &                          0.1533 \\
&           &  NP  &     0.3598  &          0.2232  &                        0.3197  & \textbf{0.1629} &     0.1030  &          0.1321  &                        0.0728  & \textbf{0.0412}\\ \hline
&  2  &  AUC  & \textbf{0.8019} &          0.7935  &                        0.8018  &                          0.7937 & \textbf{0.8409} &          0.8366  &                        0.8354  &                          0.8382 \\
&           &  WSD  &     0.1330  & \textbf{0.0188} &                        0.1052  &                          0.0310 &     0.0480  &          0.0238  &                        0.0327  & \textbf{0.0151}\\
&           &  VSD  &     0.0524  & \textbf{0.0093} &                        0.0517  &                          0.0284 &     0.2564  & \textbf{0.1308} &                        0.1958  &                          0.1434 \\
&           &  NP  &     0.3617  &          0.2082  &                        0.3232  & \textbf{0.0689} &     0.1553  &          0.0923  &                        0.1020  & \textbf{0.0530}\\ \hline
&  3  &  AUC  & \textbf{0.8307} &          0.8298  &                        0.8280  &                          0.8273 &     0.8325  &          0.8302  & \textbf{0.8336} &                          0.8303 \\
&           &  WSD  &     0.1845  &          0.0450  &                        0.1263  & \textbf{0.0246} &     0.0650  &          0.0283  &                        0.0372  & \textbf{0.0058}\\
&           &  VSD  &     0.1129  &          0.0264  &                        0.0432  & \textbf{0.0219} &     0.4439  &          0.1769  &                        0.2569  & \textbf{0.0857}\\
&           &  NP  &     0.4093  &          0.2374  &                        0.2891  & \textbf{0.0134} &     0.2052  &          0.1677  &                        0.1057  & \textbf{0.0104}\\ \hline
&  Avg  &  AUC  & \textbf{0.8106} &          0.8047  &                        0.8083  &                          0.8066 & \textbf{0.8357} &          0.8337  &                        0.8353  &                          0.8322 \\
&           &  WSD  &     0.1545  &          0.0288  &                        0.0990  & \textbf{0.0230} &     0.0387  &          0.0231  &                        0.0260  & \textbf{0.0086}\\
&           &  VSD  &     0.0887  & \textbf{0.0144} &                        0.0428  &                          0.0191 &     0.2927  &          0.1565  &                        0.1894  & \textbf{0.1216}\\
&           &  NP  &     0.3807  &          0.2158  &                        0.0990  & \textbf{0.0230} &     0.1448  &          0.1217  &                        0.0260  & \textbf{0.0086}\\ \hline\hline
Middle  &  0  &  AUC  & \textbf{0.8125} &          0.8072  &                        0.8125  &                          0.8005 & \textbf{0.8384} &          0.8372  &                        0.8210  &                          0.8384 \\
&           &  WSD  &     0.2021  & \textbf{0.0228} &                        0.1659  &                          0.1093 &     0.0457  &          0.0138  & \textbf{0.0027} &                          0.0457 \\
&           &  VSD  &     0.1399  & \textbf{0.0138} &                        0.0852  &                          0.0282 &     0.3317  &          0.1420  & \textbf{0.0769} &                          0.2913 \\
&           &  NP  &     0.4361  &          0.2050  &                        0.3823  & \textbf{0.0582} &     0.1336  &          0.0927  &                        0.0194  & \textbf{0.0168}\\ \hline
&  1  &  AUC  &     0.7888  &          0.7824  & \textbf{0.7938} &                          0.7893 & \textbf{0.8352} &          0.8320  &                        0.8336  &                          0.8289 \\
&           &  WSD  &     0.1196  &          0.0203  &                        0.0991  & \textbf{0.0064} &     0.0295  &          0.0230  &                        0.0162  & \textbf{0.0055}\\
&           &  VSD  &     0.0552  &          0.0083  &                        0.0524  & \textbf{0.0044} &     0.1818  &          0.3919  &                        0.1134  & \textbf{0.0615}\\
&           &  NP  &     0.3467  &          0.2055  &                        0.3247  & \textbf{0.1249} &     0.0624  &          0.1153  &                        0.0547  & \textbf{0.0320}\\ \hline
&  2  &  AUC  & \textbf{0.8002} &          0.7923  &                        0.7987  &                          0.7970 &     0.8351  &          0.8351  & \textbf{0.8374} &                          0.8180 \\
&           &  WSD  &     0.1147  & \textbf{0.0172} &                        0.0878  &                          0.0651 &     0.0459  &          0.0322  &                        0.0235  & \textbf{0.0104}\\
&           &  VSD  &     0.0452  & \textbf{0.0046} &                        0.0317  &                          0.0257 &     0.4267  &          0.3616  &                        0.2495  & \textbf{0.0954}\\
&           &  NP  &     0.3443  &          0.2140  &                        0.2988  & \textbf{0.1111} &     0.1478  &          0.0905  &                        0.1076  & \textbf{0.0117}\\ \hline
&  3  &  AUC  &     0.8205  &          0.8257  &                        0.8251  & \textbf{0.8344} &     0.8306  & \textbf{0.8324} &                        0.8297  &                          0.8298 \\
&           &  WSD  &     0.2641  & \textbf{0.0095} &                        0.0935  &                          0.0161 &     0.1292  &          0.0076  &                        0.0447  & \textbf{0.0020}\\
&           &  VSD  &     0.1139  & \textbf{0.0050} &                        0.0486  &                          0.0073 &     0.5266  & \textbf{0.0560} &                        0.2816  &                          0.1179 \\
&           &  NP  &     0.4549  &          0.2168  &                        0.3064  & \textbf{0.0966} &     0.2723  &          0.1373  &                        0.1365  & \textbf{0.0136}\\ 
\hline
&  Avg  &  AUC  &     0.8055  &          0.8019  & \textbf{0.8075} &                          0.8053 & \textbf{0.8348} &          0.8342  &                        0.8304  &                          0.8288 \\
&           &  WSD  &     0.1751  & \textbf{0.0175} &                        0.1116  &                          0.0492 &     0.0626  &          0.0192  &                        0.0218  & \textbf{0.0159}\\
&           &  VSD  &     0.0885  & \textbf{0.0079} &                        0.0545  &                          0.0164 &     0.3667  &          0.2379  &                        0.1804  & \textbf{0.1415}\\
&           &  NP  &     0.3955  &          0.2103  &                        0.1116  & \textbf{0.0492} &     0.1540  &          0.1089  &                        0.0218  & \textbf{0.0159}\\
\bottomrule
\end{tabular}
\label{tab:res-adults}
\end{table*}

\begin{algorithm}[ht]
\caption{$\textsc{Fairness Enhanced Data Augmentation}$ }
\begin{algorithmic}[1]
\State \textbf{Input: }$s \in \{ \text{Augment},\text{Remove}  \},X^{(i)}_{train} \in \mathbbm{R}^{n^{(i)} \times d}, $

$\bm y^{(i)}_{train} \in \{0,1\}^{n^{(i)}}, i=0$


\While{ $i \leq 20 \text{ and } 1 \leq n^{(i)} \leq 3n^{(0)} $ }
\State \textcolor{blue}{//Train model with current training data} 
\State \textcolor{blue}{//$n^{(i)}$ is the \#training data in $i^{th}$ round} 
\State $ M^{(i)} = \textsc{Train}(X^{(i)}_{train}, \bm y^{(i)}_{train})$
\State \textcolor{blue}{//Based on current trained model, process data for the next round.}
\State $ X^{(i+1)}_{train}, \bm y^{(i+1)}_{train} = \textsc{Proc}(X^{(i)}_{train}, \bm y^{(i)}_{train}, M^{(i)}, s)$
\State $i \pluseq 1 $

\EndWhile
\State \Return $M^{(i)}$

\noindent\hrulefill

\Procedure{Proc}{$X, \bm y, M, s$}
\State \textbf{Input: } $s \in \{\text{Augment},\text{Remove}  \},X \in \mathbbm{R}^{k \times d}, $

$\bm y \in \{0,1\}^{k}, M \text{ is the model }$
\State $Ind_{flip} = []$
\For{$ i \in [k]  $ }
\State $ y_i = M(\bm x_i) $
\State \textcolor{blue}{ //$x'_i$ is the feature vector after flipping the protected feature.} 
\State $ y'_i = M(\bm x'_i) $ 
\If{$y'_i \neq y_i$}
\State $Ind_{flip}.\operatorname{append}(i)$
\EndIf
\EndFor
\If{$s == \text{Remove}$ }
    \State \textcolor{blue}{//Remove the data with different results after the feature flipped}
    \State $X_{new} = \{ \bm x_j | j \in [k],  j \notin Ind_{flip} \}$ 
    \State $\bm y_{new} = \{ y_j | j \in [k],  j \notin Ind_{flip} \}$
    \State \Return $ X_{new} ,  \bm y_{new}$
\EndIf

\If{$s == \text{Augment}$ }
    \State \textcolor{blue}{//Augment the data with the feature flipped and consistent label}
    \State $X_{new} = \{ \bm x'_j | j \in Ind_{flip} \}$ 
    \State $\bm y_{new} = \{ y_j | j \in Ind_{flip} \}$
    \State $X_{new} = [X_{new}; X]$
    \State $\bm y_{new} = [\bm y_{new}; \bm y]$
    \State \Return $ X_{new} ,  \bm y_{new}$
\EndIf

\EndProcedure
\end{algorithmic}
\label{algo:fair-data-aug}
\end{algorithm}

\begin{figure*}
    \centering
    \includegraphics[page=1,width=0.95\textwidth]{figs/fig-compas-fairness-vs-auc-scatter-diff.pdf}
    \includegraphics[page=2,width=0.95\textwidth]{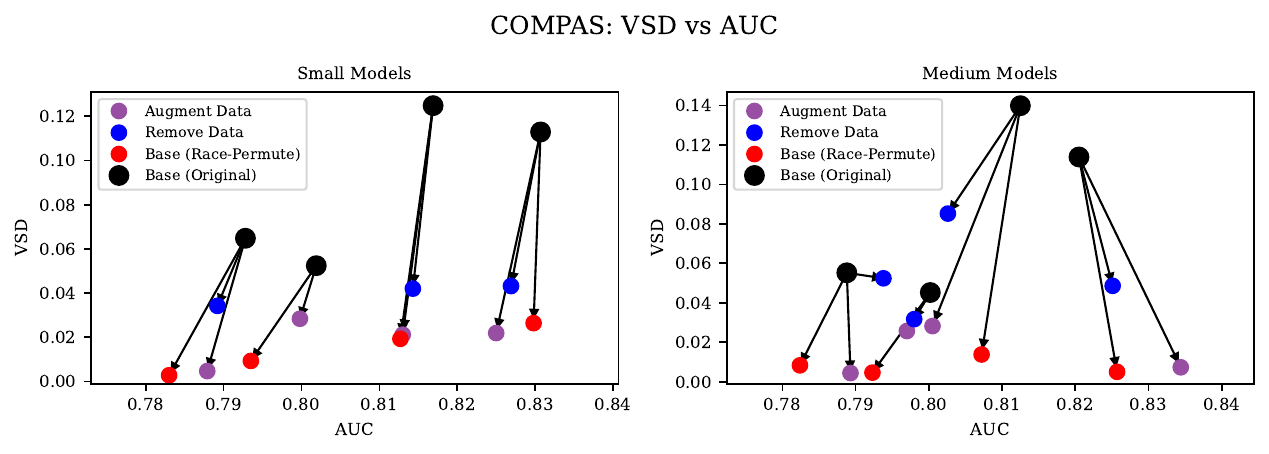}
    \includegraphics[page=3,width=0.95\textwidth]{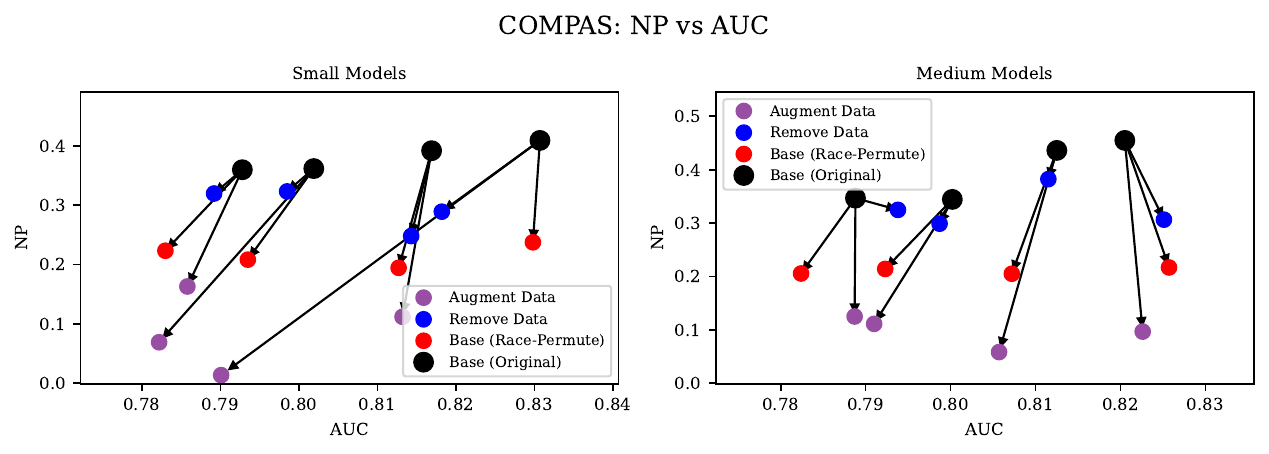}

    \caption{COMPAS Dataset: Model performance/AUC vs Model fairness/NP under three fairness-sensitive training methods.  Arrows down and to the left reflect models that are fairer but less accurate than the original mode.  We obtain large improvements in fairness at little cost in accuracy. }
    \label{fig:auc-vs-fairness-compas-long}
\end{figure*}

\begin{figure*}
    \centering
    \includegraphics[page=1,width=0.95\textwidth]{figs/fig-adults-fairness-vs-auc-scatter-diff.pdf}
    \includegraphics[page=2,width=0.95\textwidth]{figs/fig-adults-fairness-vs-auc-scatter-diff.pdf}
    \includegraphics[page=3,width=0.95\textwidth]{figs/fig-adults-fairness-vs-auc-scatter-diff.pdf}
    \caption{ADULTS Dataset: Model performance/AUC vs Model fairness/NP under three fairness-sensitive training methods.  Arrows down and to the left reflect models that are fairer but less accurate than the original mode.  We obtain large improvements in fairness at little cost in accuracy. }
    \label{fig:auc-vs-fairness-adults-long}
\end{figure*}

\end{document}